\documentclass[11pt]{article}
\usepackage[margin=1in]{geometry}
\usepackage[T1]{fontenc}
\usepackage[utf8]{inputenc}
\usepackage{textcomp}
\usepackage{amsmath,amssymb}
\usepackage{graphicx}
\usepackage{xcolor}
\usepackage{hyperref}
\hypersetup{hidelinks}
\title{New Benchmarking Shows Limited Generalization Power of TCR Antigenic Epitope Prediction Models}

\author{\begin{minipage}{0.96\textwidth}\centering
Yiming Liao\textsuperscript{1},
Yiheng Li\textsuperscript{2},
Ning Jiang\textsuperscript{3,4,5,6,7,8,*},
Bo Li\textsuperscript{2,*},
Keke Chen\textsuperscript{1,*}\\[0.75em]
\small \textsuperscript{1}Trustworthy and Intelligent Computing Lab (TAIC), Department of Computer Science and Electrical Engineering, University of Maryland, Baltimore County, Baltimore, Maryland 21250, USA\\
\small \textsuperscript{2}Children's Hospital of Philadelphia, Philadelphia, PA 19104, USA\\
\small \textsuperscript{3}Department of Bioengineering, University of Pennsylvania, Philadelphia, PA 19104, USA\\
\small \textsuperscript{4}Institute for Immunology \& Immune Health, University of Pennsylvania, Philadelphia, PA 19104, USA\\
\small \textsuperscript{5}Institute for RNA Innovation, University of Pennsylvania, Philadelphia, PA 19104, USA\\
\small \textsuperscript{6}Abramson Cancer Center, University of Pennsylvania, Philadelphia, PA 19104, USA\\
\small \textsuperscript{7}Center for Precision Engineering for Health, University of Pennsylvania, Philadelphia, PA 19104, USA\\
\small \textsuperscript{8}Center for Cellular Immunotherapies, University of Pennsylvania, Philadelphia, PA 19104, USA\\
\small \textsuperscript{*}Corresponding authors.
\end{minipage}}

\date{}

\begin{document}
\maketitle

\begin{abstract}
Accurate computational prediction of T cell receptor (TCR) antigen specificity would transform the study of T cell biology and enable scalable immune engineering, yet existing models lack sufficient sensitivity and specificity for broad applications. A major limitation is the absence of rigorously defined, unseen benchmark datasets that allow unbiased evaluation of model performance and generalizability. Here, we describe two complementary classes of datasets that meet this criterion and argue that they provide both a robust framework for model assessment and a foundation for next-generation TCR--antigen prediction algorithm development.
\end{abstract}

\section*{Benchmark Design and Results}
T cell receptors (TCRs) are essential to adaptive immunity, mediating
the recognition of antigenic epitopes presented by major
histocompatibility complexes (MHCs). In the past 10 years, we have
witnessed an exponential increase in the development and application of
various models to computationally predict TCR antigenic epitopes.
However, recent studies (Nielsen, Eugster et al. 2024; Lu, Wang et al.
2026) evaluating these models concluded that they poorly predict TCR
antigen specificity for truly unseen data. Both studies emphasized the
importance of training dataset size, the source of negative TCRs in the
training data, and the proper use of independent test sets for
performance evaluation. After extensive comparisons, both studies raised
concerns about the ability of these models to generalize TCR antigen
prediction to unseen antigens. As the field continues to develop deep
learning models, identifying unseen data from published experimental
datasets for unbiased model evaluation and future model learning has
become a critical challenge.

To address this challenge, we carefully analyzed caveats in current
model evaluation practice. Model developers often use datasets from
different studies for training and testing. However, many TCR antigen
discovery papers use common viral antigens as controls. As a result,
similar antigens or TCRs repeatedly appear across studies, confounding
model evaluation that is intended to use unseen data. We therefore
created a benchmarking framework using a recent publicly available
high-throughput TCR-antigen mapping dataset that has not been uploaded
to the two commonly used databases, VDJdb and IEDB. This dataset comes
from a high-dimensional, high-throughput single antigen-specific T cell
profiling study of millions of T cells binding to hundreds of
DNA-barcoded viral and type 1 diabetes (T1D) peptide MHC (pMHC)
tetramers (TetTCR-SeqHD) (Ma, Schonnesen et al. 2021).

Although this dataset has not been used to benchmark TCR-epitope binding
prediction and could therefore be considered a new dataset, it
contains TCRs targeting 17 epitopes found in public databases and
therefore represents ``seen'' data. Thus, a key aspect of our framework
involved removing TCRs \textbf{and antigens} from the TetTCR-SeqHD dataset
if they overlap with any training dataset used by the eight models
tested, including ERGO, ERGO-II, NetTCR2.0, NetTCR2.2, TITAN, PanPep,
SCEPTR, and EPACT.

We further reasoned that because the TetTCR-SeqHD study used two groups
of antigens--epitopes from common viruses with many cognate TCRs
reported in public databases and epitopes from T1D-associated
self-proteins with few reported cognate TCRs--a more general approach
would be to separate antigens and associated TCRs by organism or tissue
origin and test models on these antigen groups separately. Epitopes
derived from common viruses or bacteria would be expected to receive
higher cognate-TCR prediction scores, whereas epitopes derived from
self antigens or cancer neoantigens would be expected to receive lower
scores.

To test this hypothesis, we implemented a multi-step benchmarking
process that included dataset preprocessing, removing duplicate TCRs
from the training data, generating negative samples by swapping positive
epitopes and TCR pairs, prediction inference, post-processing grouping
(such as selecting unseen and seen groups or viral and T1D groups), and
result evaluation (Fig. 1a). This process removed duplicate or
nearly identical TCR/\textbf{antigenic-epitope} pairs from benchmark
test datasets relative to training data to prevent data leakage.
In one case, the stringent removal of duplicate or similar
sequences---allowing up to three amino acid substitutions in the CDR3$\beta$
region---eliminated between 40\% and 70\% of sequences in test datasets
depending on datasets and models. These sequences, if retained, could
artificially inflate performance through memorization.

For the TetTCR-SeqHD dataset, negatives were generated using a
controlled shuffling method: each positive TCR--epitope pair was matched
with up to five epitope-mismatched negatives that had a Levenshtein
distance greater than 3 from any existing data on TCR CDR3$\beta$ sequence.
For the IMMREP23 dataset, which already includes negative data, no
additional negatives were added. This dataset contains lab-confirmed
true negatives---cases that have been tested and shown not to react.
These examples are more reliable than ``negative'' examples created by
simply swapping components of known positives, which may still involve
unknown cross-reactions. A randomly swapped peptide may still have
biological binding affinity to the TCR in vivo, inadvertently
introducing false negatives into the evaluation set.

Performance was evaluated using the Macro AUC\textsubscript{0.1}
metric---an area under the receiver operating characteristic (ROC) curve
up to a 10\% false positive rate---selected for its sensitivity in
low-false-positive-rate (FPR) regions, which are critical for TCR
specificity evaluation. Results were averaged across epitope groups and
grouped by viral versus self epitopes for the TetTCR-SeqHD dataset and
seen versus unseen epitopes for the IMMREP23 dataset (Fig. 1b). Unseen
epitopes in IMMREP23 (e.g., SALPTNADLY, TSDACMM) are strictly defined as
peptides that are completely uncharacterized or lack prior
representation in public TCR databases\textsuperscript{10} (such as
VDJdb or IEDB). Consistent with our hypothesis, in general, most models
achieved moderate AUC$_{0.1}$ scores (\textasciitilde0.5--0.67) on ``Viral''
or ``Seen'' epitopes, with NetTCR-2.2 reaching the highest performance
on IMMREP23 ``Seen'' group (AUC$_{0.1}$ = 0.600). However, performance on
``Self'' or ``Unseen'' epitopes consistently dropped to near-random
levels (Fig. 1b). In both the ``Viral'' and ``Seen'' groups, a few
epitopes had associated TCRs with AUC\textsubscript{0.1} scores above 0.6.
Upon further examination, these are common epitopes derived from HIV,
HCV, Flu, and EBV.

Because high-throughput TCR and antigenic-epitope pairing datasets
remain limited, we explored the feasibility of using another type of
data to evaluate TCR-antigen prediction models: TCR pMHC tetramer
staining with peptide mutants. This Fingerprinting dataset is easier to
obtain and provides more quantitative binding measurements. We used our
recent dataset (Malone, Huang et al. 2025), which screened 21 TCRs for
binding to 172 possible single amino acid mutations of the dominant
SARS-CoV-2 epitope YLQPRTFLL (YLQ). Of these 21 TCRs, 20 were
identified using the YLQ wild-type antigen, while TCR21 was identified
using a pool of 19 R5 mutants. This TCR has a distinct binding pattern
compared with the other 20 YLQ wild-type-specific TCRs, as shown by the
pMHC tetramer binding heatmap (Fig. 1f). Thus, this represents a unique set of
quantitative cross-reactive TCR-pMHC interactions.

Our analysis showed that the performance of the eight models evaluated
is indistinguishable from random assignment (Fig. 1c). The quantitative
nature of the Fingerprinting dataset allowed us to further dissect
individual model performance. Spearman's correlation
analysis showed no statistically significant correlations for any
model. As an alternative to this global correlation analysis, we also
correlated Fingerprinting data with model predictions for each TCR. This
analysis showed that three models, ERGO2, NetTCR2.2, and EPACT, had some
positive correlations (Fig. 1e). However, further comparison of heatmaps
from Fingerprinting experiments and model predictions revealed that both
the overall pattern and the degree of binding did not match well at the
individual TCR level (Fig. 1f--i).

\begin{figure}[p]
\centering
\includegraphics[width=\textwidth,height=0.62\textheight,keepaspectratio]{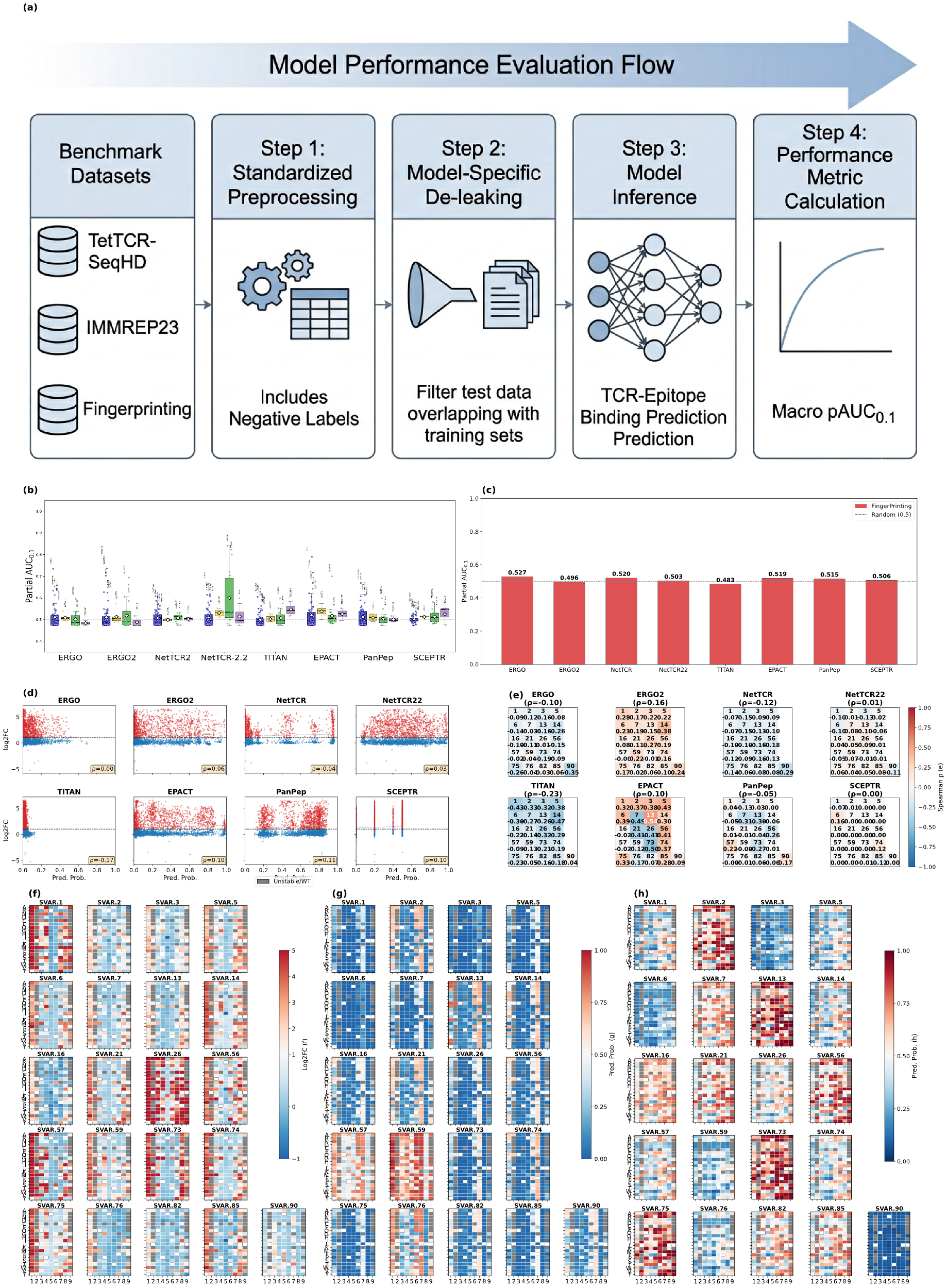}
\caption{\textbf{Systematic benchmarking of TCR--peptide binding models across diverse epitope landscapes.}
\textbf{(a)} \textbf{Benchmark pipeline overview.} Schematic of the evaluation workflow: three datasets (\textbf{TetTCR-SeqHD}, \textbf{IMMREP23}, \textbf{Fingerprinting}) undergo standardized preprocessing and model-specific data-leakage control before inference. Performance is quantified using macro-averaged partial AUC (pAUC\textsubscript{0.1}).
\textbf{(b)} \textbf{Performance on viral/self (TetTCR-SeqHD) and seen/unseen (IMMREP) epitopes.} Distribution of per-peptide pAUC\textsubscript{0.1} scores for the TetTCR-SeqHD and IMMREP23 datasets. Boxplots represent the interquartile range (IQR) across individual peptides (points).
\textbf{(c)} \textbf{Model performance on Fingerprinting peptides.} Comparison of model pAUC\textsubscript{0.1} on the Fingerprinting dataset; the dashed line indicates random performance (0.5).
\textbf{(d)} \textbf{Correlation between predicted probability and experimental enrichment.} Scatter plots showing predicted binding probability versus experimental log\textsubscript{2} fold-change (FC) for Fingerprinting dataset variants. Points are colored by binding status; $\rho$ indicates the Spearman correlation.
\textbf{(e)} \textbf{TCR-specific predictive accuracy.} Heatmaps displaying Spearman correlations between predictions and log\textsubscript{2} FC for individual TCRs across all models.
\textbf{(f--h)} \textbf{Comparative binding fingerprints.} \textbf{(f)} Experimental log\textsubscript{2} FC landscape across peptide variants (positions 1--9) for 17 SVAR TCRs. Corresponding predicted binding fingerprints of probabilities and log\textsubscript{2} FC correlations are shown for \textbf{(g)} ERGO2 and \textbf{(h)} NetTCR-2.2.}
\label{fig:benchmark}
\end{figure}

\section*{Discussion}

One of the most critical insights from this study is that performance
degradation is more influenced by epitope than by TCR novelty. In
contrast to prior assumptions, we find that most models have implicitly
memorized epitope-specific features rather than learning generalizable
binding rules. This was particularly evident when comparing model
predictions on epitope-unseen versus TCR-unseen conditions. To
rigorously evaluate model generalization, we introduced the TetTCR-SeqHD
and Fingerprinting datasets, which were not available in public
databases such as VDJdb or IEDB, to ensure truly unseen data for
unbiased evaluation. When only the TCR was novel, performance remained
stable. In contrast, YLQ R5 substitutions represent truly unseen epitope
bindings and consistently reduced predictive power across all models
tested. This limitation in generalization ability is a critical
bottleneck that the field must address before these models can be
reliably deployed in clinical and translational applications. The lack
of generalizability is likely driven by uneven data quality, an
extremely skewed TCR-to-epitope ratio, and limited training data relative
to the immense antigen space. These results suggest that current
TCR-epitope prediction models may overfit to the limited epitope
diversity present in training data, raising serious concerns about their
utility for predicting responses to novel pathogens or neoantigens.

Current TCR-epitope binding databases are heavily skewed toward a small
number of well-studied viral epitopes, while the vast majority of
antigenic space remains unexplored. Models trained predominantly on
viral epitopes may not generalize to tumor neoantigens, autoantigens, or
emerging pathogen epitopes. Our results suggest that
simply expanding the training corpus with more seen examples is
insufficient; instead, model architecture, training strategies, and
feature representations must evolve to capture higher-order
immunological principles. Future model development must prioritize
balanced training data with adequate representation of non-viral
antigens, requiring coordinated efforts to generate experimental data
across underrepresented epitope categories.

The quantitative, systematic nature of the Fingerprinting dataset
represents a valuable opportunity not only for rigorous model testing
but also for training next-generation models capable of predicting
cross-reactive TCR binding. Current models showed no statistically
significant correlations with experimental binding data on this dataset.
However, the structured variation, in which each position is
systematically mutated, provides rich information about binding
landscapes that could be leveraged for model training. Models could
learn nuanced cross-reactivity patterns, understanding quantitative
binding affinity spectra across related epitope variants rather than
binary classifications. This capability is critical for predicting
responses to viral escape mutants, identifying off-target reactivities
in TCR therapeutics, and understanding autoimmune cross-reactivity.

In conclusion, this work introduces a rigorous and reproducible
benchmark that reveals current deep learning models fall short of the
performance needed for generalization to unseen TCR--epitope pairs. Our
results suggest that the development of robust pan-specific prediction
frameworks remains an open and pressing problem. This benchmark should
serve as both a call to action and a resource for the community,
emphasizing the need for realistic negative sampling, stringent
deduplication, and functional definitions of epitope novelty in future
studies.

\section*{Author Contributions}

Y. Liao designed the benchmark experiments. Y. Liao and Y. Li cleaned
the data and performed the experiments. N. Jiang provided the
TetTCR-SeqHD and Fingerprinting datasets. B. Li, K. Chen, and N. Jiang
analyzed the results. Y. Liao and K. Chen prepared the manuscript with
input from all authors.

\section*{Competing Interests}

The authors declare no competing interests.

\section*{References}

1. Hedrick, S.M., Cohen, D.I., Nielsen, E.A. \& Davis, M.M. Isolation of
cDNA clones encoding T cell-specific membrane-associated proteins.
\emph{Nature} \textbf{308}, 149-153 (1984).

2. Yanagi, Y. et al. A human T cell-specific cDNA clone encodes a
protein having extensive homology to immunoglobulin chains.
\emph{Nature} \textbf{308}, 145-149 (1984).

3. Bevan, M.J. The major histocompatibility complex determines
susceptibility to cytotoxic T cells directed against minor
histocompatibility antigens. \emph{J Exp Med} \textbf{142}, 1349-1364
(1975).

4. Gordon, R.D., Simpson, E. \& Samelson, L.E. In vitro cell-mediated
immune responses to the male specific(H-Y) antigen in mice. \emph{J Exp
Med} \textbf{142}, 1108-1120 (1975).

5. Kappler, J.W. \& Marrack, P.C. Helper T cells recognise antigen and
macrophage surface components simultaneously. \emph{Nature}
\textbf{262}, 797-799 (1976).

6. Bjorkman, P.J. et al. The foreign antigen binding site and T cell
recognition regions of class I histocompatibility antigens.
\emph{Nature} \textbf{329}, 512-518 (1987).

7. Garcia, K.C. et al. An alphabeta T cell receptor structure at 2.5 A
and its orientation in the TCR-MHC complex. \emph{Science} \textbf{274},
209-219 (1996).

8. Ma, K.-Y. et al. High-throughput and high-dimensional single-cell
analysis of antigen-specific CD8+ T cells. \emph{Nature Immunology}
\textbf{22}, 1590-1598 (2021).

9. Malone, M.J. et al. Resistance Potential of the HLA-A2-restricted
Immunodominant SARS-CoV-2 Specific CD8+ T Cell Receptor Repertoire to
Antigenic Drift. \emph{Nature Communications}, Accepted.

10. Nielsen, M. et al. Lessons learned from the IMMREP23 TCR-epitope
prediction challenge. \emph{ImmunoInformatics} \textbf{16}, 100045
(2024).

Lu, Y., Y. Wang, M. Xu, B. Xie, Y. Yang, H. Xu and S. Suo (2026).
"Assessment of computational methods in predicting TCR-epitope binding
recognition." \emph{Nat Methods} \textbf{23}(1): 248-259.

Ma, K.-Y., A. A. Schonnesen, C. He, A. Y. Xia, E. Sun, E. Chen, K. R.
Sebastian, Y.-W. Guo, R. Balderas, M. Kulkarni-Date and N. Jiang (2021).
"High-throughput and high-dimensional single-cell analysis of
antigen-specific CD8+ T cells." \emph{Nature Immunology}
\textbf{22}(12): 1590-1598.

Malone, M. J., C. Huang, Y. Zhang, Y. Qi, L. Williams, L. F. Su, J. Lou
and N. Jiang (2025). "Resistance potential of the HLA-A2-restricted
immunodominant SARS-CoV-2-specific CD8(+) T cell receptor repertoire to
antigenic drift." \emph{Nat Commun}.

Nielsen, M., A. Eugster, M. F. Jensen, M. Goel, A. Tiffeau-Mayer, A.
Pelissier, S. Valkiers, M. R. Martínez, B. Meynard-Piganeeau and V.
Greiff (2024). "Lessons learned from the IMMREP23 TCR-epitope prediction
challenge." \emph{ImmunoInformatics} \textbf{16}: 100045.

\end{document}